\def\year{2021}\relax
\documentclass[letterpaper]{article} % DO NOT CHANGE THIS
\usepackage{aaai21}  % DO NOT CHANGE THIS
\usepackage{times}  % DO NOT CHANGE THIS
\usepackage{helvet} % DO NOT CHANGE THIS
\usepackage{courier}  % DO NOT CHANGE THIS
\usepackage[hyphens]{url}  % DO NOT CHANGE THIS
\usepackage{graphicx} % DO NOT CHANGE THIS
\usepackage{amsfonts,amssymb}
\usepackage{amsmath}
\usepackage{multirow}
\usepackage{blkarray}
\usepackage[switch]{lineno}
\usepackage{bbm}
\usepackage{natbib}  % DO NOT CHANGE THIS AND DO NOT ADD ANY OPTIONS TO IT
\usepackage{caption} % DO NOT CHANGE THIS AND DO NOT ADD ANY OPTIONS TO IT
\title{Joint Entity and Relation Extraction with Set Prediction Networks}
\author {
        Dianbo Sui\textsuperscript{\rm $\heartsuit$ $\spadesuit$ }
        Yubo Chen\textsuperscript{\rm $\heartsuit$}
        Kang Liu\textsuperscript{\rm $\heartsuit$ $\spadesuit$ } 
        Jun Zhao\textsuperscript{\rm $\heartsuit$ $\spadesuit$ }  \\
        Xiangrong Zeng \textsuperscript{\rm $\diamondsuit$} Shengping Liu \textsuperscript{\rm $\diamondsuit$}\\
}
\affiliations {
    \textsuperscript{\rm $\heartsuit$} National Laboratory of Pattern Recognition, Institute of Automation, Chinese Academy of Sciences, Beijing, China\\
	 \textsuperscript{\rm $\spadesuit$} University of Chinese Academy of Sciences, Beijing, China\\
    \textsuperscript{\rm $\diamondsuit$} Beijing Unisound Information Technology Co., Ltd, Beijing, China\\
    \{dianbo.sui, yubo.chen, kliu,  jzhao\}@nlpr.ia.ac.cn, 
	\{zengxiangrong, liushengping\}@unisound.com
}
\begin{document}
\maketitle

\begin{abstract}
The joint entity and relation extraction task aims to extract all relational triples from a sentence. In essence, the relational triples contained in a sentence are unordered. However, previous seq2seq based models require to convert the set of triples into a sequence in the training phase. To break this bottleneck, we treat joint entity and relation extraction as a direct set prediction problem, so that the extraction model can get rid of the burden of predicting the order of multiple triples. To solve this set prediction problem, we propose networks featured by transformers with non-autoregressive parallel decoding. Unlike autoregressive approaches that generate triples one by one in a certain order, the proposed networks directly output the final set of triples in one shot. Furthermore, we also design a set-based loss that forces unique predictions via bipartite matching. Compared with cross-entropy loss that highly penalizes small shifts in triple order, the proposed bipartite matching loss is invariant to any permutation of predictions; thus, it can provide the proposed networks with a more accurate training signal by ignoring triple order and focusing on relation types and entities. Experiments on two benchmark datasets show that our proposed model significantly outperforms current state-of-the-art methods. Training code and trained models will be available at \url{http://github.com/DianboWork/SPN4RE}.
\end{abstract}

\section{Introduction}

A relational triple consists of two entities connected by a semantic relation, which is in the form of (subject, relation, object). The extraction of relational triples from unstructured raw texts is a key technology for automatic knowledge graph construction, which has received growing interest in recent years.

There have been several studies addressing technical solutions for relational triple extraction. Early researches, such as \citet{zelenko2003kernel,chan2011exploiting}, employ a pipeline manner to extract both of entities and relations, where entities are recognized first and then the relation between the extracted entities is predicted. Such a pipeline approach ignores the relevance of entity identification and relation prediction \cite{li2014incremental} and tends to suffer from the error propagation problem. 
% \cite{zhang2017end} 

To model cross-task dependencies explicitly and prevent error propagation in the pipeline approach, subsequent studies propose joint entity and relation extraction. These studies can be roughly categorized into three main
paradigms. The first stream of work, such as \citet{miwa2016end,gupta2016table,zhang2017end}, treats joint entity and relation extraction task as an end-to-end table filling problem. Although these methods represent entities and relations with shared parameters in a single model, they extract the entities and relations separately and produce redundant information \cite{zheng2017joint}. The second stream of work, such as \citet{zheng2017joint,dai2019joint,wei-etal-2020-novel}, transforms joint entity and relation extraction into sequence labeling. To do this, human experts need to design a complex tagging schema. The last stream of work, including \citet{zeng2018extracting,zeng2019learning,nayak2019ptrnetdecoding,zeng2020copymtl}, is driven by the sequence-to-sequence (seq2seq) model \cite{sutskever2014sequence} to generate relational triples directly, which is a flexible framework to handle overlapping triples and does not require the substantial effort of human experts.

We follow the seq2seq based models for joint entity and relation extraction. Despite the success of existing seq2seq based models, they are still limited by the autoregressive decoder and the cross-entropy loss. The reasons are as follows: the relational triples contained in a sentence have no intrinsic order in essence. However, in order to adapt the autoregressive decoder, whose output is a sequence,  the unordered target triples must be sorted in a certain order during the training phase. Meanwhile, cross-entropy is a permutation-sensitive loss function, where a penalty is incurred for every triple that is predicted out of the position. Consequently, current seq2seq base models not only need to learn how to generate triples, but also are required to consider the extraction order of multiple triples. 

% consists of three parts  featured by transformers with non-autoregressive parallel decoding and the bipartite matching loss.  In detail, there are three parts in the proposed set prediction networks (SPN):  to avoid introducing the order of triplets 
% restoring to the original form of this task without considering the order of multiple triples
In this work, we formulate the joint entity and relation extraction task as a set prediction problem, avoiding considering the order of multiple triples. In order to solve the set prediction problem, we propose an end-to-end network featured by transformers with non-autoregressive parallel decoding and bipartite matching loss. In detail, there are three parts in the proposed set prediction networks (SPN): a sentence encoder, a set generator, and a set based loss function. First of all, we adopt the BERT model \cite{devlin2018bert} as the encoder to represent the sentence. Then, since an autoregressive decoder must generate items one by one in order, such a decoder is not suitable for generating unordered sets. In contrast, we leverage the transformer-based non-autoregressive decoder \cite{gu2018non} as the set generator, which can predict all triples at once and avoid sorting triples. Finally, in order to assign a predicted triple to a unique ground truth triple, we propose bipartite matching loss function inspired by the assigning problem in operation research \cite{kuhn1955hungarian,munkres1957algorithms,edmonds1972theoretical}. Compared with  cross-entropy loss 
that highly penalizes small shifts in 
triple order, the proposed loss function is invariant to any permutation of predictions; thus it is suitable for evaluating the difference between ground truth set and prediction set.

% To summarize, our contributions are as follows:
In a nutshell, our main contributions are:
% the main contributions of our work are as follows:

\begin{itemize}
\item We formulate the joint entity and relation extraction task as a set prediction problem.
\item We combine non-autoregressive parallel decoding with bipartite matching loss function to solve this problem.
\item Our proposed method yields state-of-the-art results on two benchmark datasets, and we perform various experiments to verify the effectiveness of the method.
\textbf{}
\end{itemize}
% the conjunction of the bipartite matching loss and transformers with
% (non-autoregressive) parallel decoding 
% Our work build on prior work in several domains:relation extraction, non-autoregressive model, andbipartite matching losses for set prediction.
% \textbf{Relation Extraction. }  \textbf{Non-autoregressive Model. } 
\section{Related Work}
% Our work builds on prior work in relation extraction and non-autoregressive model.
\subsection{Relation Extraction}
Relation extraction is a long-standing natural language process task of mining factual knowledge from free texts. When giving a sentence with annotated entities, this task degenerates into a simple task, namely relation classification. Some studies, such as \citet{zeng-etal-2014-relation,xu2015classifying}, leveraged CNN or RNN to solve the relation classification task. However, these methods ignore the extraction of entities from sentences and could not truly extract relational facts. 

When giving a sentence without any annotated entities, researchers proposed several methods to extract entities and relations jointly. Existing studies on multiple relation extraction
task can be divided into four paradigms: (1) Pipeline based methods, such as \citet{zelenko2003kernel,chan2011exploiting}, firstly recognize entities and then conduct relation classification; (2) Table filling based methods, like \citet{miwa2016end,gupta2016table,zhang2017end}, represent entities and relations with shared parameters, but extract the entities and relations separately; (3) Tagging based methods, such as \citet{zheng2017joint,dai2019joint,wei-etal-2020-novel}, treat this task as a sequence labeling problem and need to design complex tagging schema; (4) Seq2seq based methods, like \citet{zeng2018extracting,zeng2019learning,nayak2019ptrnetdecoding,zeng2020copymtl}, 
apply seq2seq model to generate relational triples directly. Our work is in line with seq2seq based methods.  In contrast with the previous studies, we reckon the triples in a sentence are in the form of a set instead of a sequence, and treat the joint entity and relation extraction as a set prediction problem.

\subsection{Non-Autoregressive Model}
Non-autoregressive models \cite{gu2018non,lee2018deterministic, ma-etal-2019-flowseq} generate all the tokens of a target in parallel and can speed up inference. Non-autoregressive models are widely explored in natural language and speech processing tasks such as neural machine translation \cite{gu2018non} and automatic speech recognition \cite{chen2019non}. To the best of our knowledge, this is the first work to apply non-autoregressive models to information extraction. In this work, we resort to the 
non-autoregressive model to generate the set of triples in one shot.

% \subsection{Set Prediction}
% There is no canonical deep learning model to directly predict sets. For constant-size set prediction, dense fully connected networks  are sufficient but costly. A general approach is to use autoregressive sequence models such as recurrent neural networks \cite{vinyals2015order}. In all cases, the loss function should be invariant by a permutation of the predictions. The usual solution is to design a loss based on the Hungarian algorithm [20], to find a bipartite matching between ground-truth and prediction. This enforces permutation-invariance, and guarantees that each target element has a unique match. We follow the bipartite matching loss approach. In contrast to most prior work however, we step away from autoregressive models and use transformers with parallel decoding, which we describe below.

\begin{figure*}[t]
  	\begin{center} \includegraphics*[clip=true,width=0.96\textwidth,height=0.3\textheight]{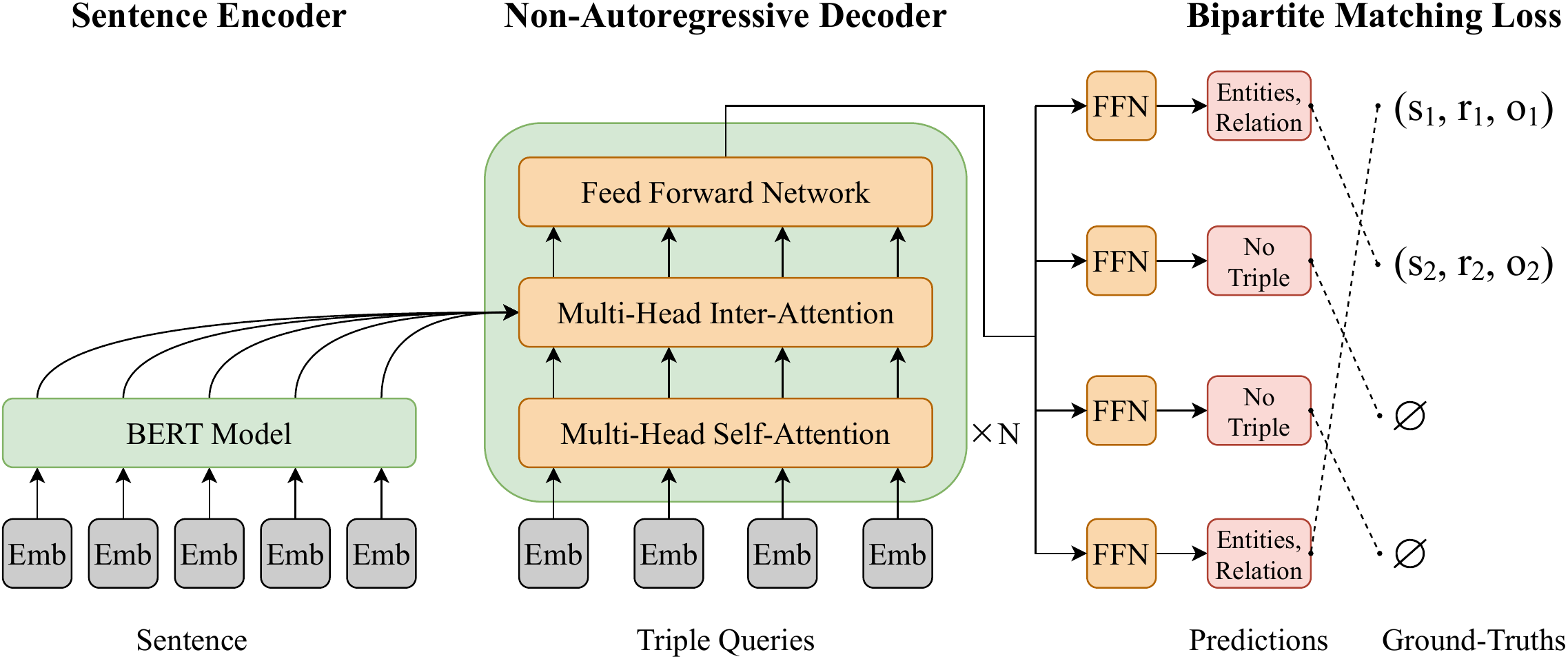}
  	\caption{The main architecture of set prediction networks.
  	The set prediction networks predict the final set of triples in parallel by combining a BERT encoder with a non-autoregressive decoder. In the training phrase, bipartite matching uniquely assigns predictions with ground truths to provide accurate training signals.
  	} \label{fig}
  \end{center}
  \end{figure*}
\section{Method}

The goal of joint entity and relation extraction is to identify all possible relational triples in a given sentence. Formally, given an raw sentence $X$, the conditional probability of the target triple set $Y=\{(s_1, r_1, o_1),..., (s_n, r_n, o_n)\}$ is: 
\begin{equation}
    P(Y|X;\theta) = p_L(n|X)\prod \limits_{i=1}^n p(Y_i|X, Y_{j \neq i}; \theta)
\label{equ1}
\end{equation}
where $p_L(n|X)$ model the size of the target triple set, and $p(Y_i|X, Y_{j \neq i}; \theta)$ means that a target triple $Y_i$ is related not only to the given sentence $X$, but also to the other triples $Y_{j \neq i}$.

In this paper, this conditional probability is parameterized using set prediction networks (SPN), which are shown in Figure \ref{fig}. Three key components of the proposed networks will be elaborated in the following section. Concretely, we first introduce the sentence encoder, which represents each token in a given sentence based on its bidirectional context. Then, we present how to use the non-autoregressive decoder to generate a set of triples in a single pass. Finally, we describe a set-based loss, dubbed as bipartite matching loss,  which forces unique matching between predicted and ground truth triples.
% Three ingredients are essential for direct set predictions in joint entity and relation extraction: (1) an encoder that represents each token in a given sentence  based on its bidirectional context; (2) a decoder that predicts a set of triplets in a single pass; (3) a set prediction loss that forces unique matching between predicted and ground truth triplets.
% We describe our architecture in detail in Figure 2.

% To solve this task, we propose set prediction networks (SPN). The overall of the networks is shown in Figure 2 and the key components of the networks will be elaborated in the following
% section. Concretely, we first introduce the sentence encoder in Section \ref{encoder}, which represents each token in a given sentence based on its bidirectional context. Then we present how to use the non-autoregressive decoder to generate a set of triplets in a single pass in Section \ref{decoder}. 

\subsection{Sentence Encoder}
\label{encoder}
The goal of this component is to obtain the context-aware representation of each token in a sentence. Given the impressive performance of recent deep transformers \cite{vaswani2017attention} trained on variants of language modeling, we utilize the BERT model \cite{devlin2018bert} as the sentence encoder.
The input sentence is segmented with tokens by the byte pair encoding \cite{sennrich-etal-2016-neural}, and then fed into the BERT model. The output of the BERT model is the context-aware embedding of tokens, and is denoted as $\mathbf{H}_e \in \mathbb{R}^{l \times d}$, where $l$ is the sentence length (including ${\rm [CLS]}$ and ${\rm [SEP]}$, two special start and end markers), and $d$ is the number of hidden units in the BERT model.

\subsection{Non-Autoregressive Decoder for Triple Set Generation}
\label{decoder}
We regard joint entity and relation extraction as a set prediction problem and use the transformer-based non-autoregressive decoder \cite{gu2018non} to directly generate the triple set. Previous studies \cite{zeng2018extracting, zeng2019learning,nayak2019ptrnetdecoding,zeng2020copymtl} transform the triple set to a triple sequence and then leverage the autoregressive decoder to generate triples one by one. In such a way, the conditional probability of the target triple set in Equation \ref{equ1} is modified into:
\begin{equation}
        P(Y|X;\theta) =\prod \limits_{i=1}^n p(Y_i|X, Y_{j < i}; \theta)
\end{equation}
In contrast, we use the non-autoregressive decoder to direct model the Equation \ref{equ1}. Compared with previous seq2seq based method, the non-autoregressive decoder can not only 
avoid learning the extraction order of multiple triples, but also generate triples based on bidirectional information, not just left-to-right information.

\textbf{Input.} Before decoding starts, the decoder need to know  the size of  the target set, in other words, $p_L(n|X)$ in Equation \ref{equ1} is required to be modeled at first. In this work, we simplify the $p_L(n|X)$ into a constant by requiring the non-autoregressive decoder to generate a fixed-size set of $m$ predictions for each sentence,  where $m$ is set to be significantly larger than the typical number of triples in a sentence.  Instead of copying tokens from the encoder side \cite{gu2018non}, the input of the decoder is initialized by $m$ learnable embeddings that we refer to as triple queries. Note that all sentences share the same triple queries.

\textbf{Architecture.} The non-autoregressive decoder is composed of a stack of $N$ identical transformer layers. In each transformer layer, there are multi-head self attention mechanism to model the relationship between triples, and multi-head inter attention mechanism to fuse the information of the given sentence. Notably, compared with autoregressive decoder, the non-autoregressive decoder does not have the constraint of an autoregressive factorization of the output, so there is no need to prevent earlier decoding steps from accessing information from later steps. Thus, there is no casual mask used in the multi-head self attention mechanism. Instead, we use the unmasked self-attention.

The $m$ triple queries are transformed into $m$ output embeddings by the non-autoregressive decoder, which are denoted as $\mathbf{H}_d \in \mathbb{R}^{m \times d}$. The output embeddings $\mathbf{H}_d$ are then independently decoded into relation types and entities by feed forward networks (FFN), resulting $m$ final predicted triples. Concretely, given an output embedding $\mathbf{h}_d \in \mathbb{R}^d$ in $\mathbf{H}_d$, the predicted relation type is obtained by:
\begin{equation}
    \mathbf{p}^r = \rm{softmax}(\mathbf{W_r} \mathbf{h}_d)
    \label{pr}
\end{equation}
and the predicted entities (subject and object) are decoded by separately predicting the starting and ending indices with four $l$-class classifiers:
\begin{align}
\mathbf{p}^{s-start} = \rm{softmax}(\mathbf{v_1^T}\rm{tanh}(\mathbf{W_1} \mathbf{h}_d + \mathbf{W_2}\mathbf{H_e}))\\
\mathbf{p}^{s-end}  = \rm{softmax}(\mathbf{v_2^T}\rm{tanh}(\mathbf{W_3} \mathbf{h}_d + \mathbf{W_4}\mathbf{H_e}))\\
\mathbf{p}^{o-start} = \rm{softmax}(\mathbf{v_3^T}\rm{tanh}(\mathbf{W_5} \mathbf{h}_d + \mathbf{W_6}\mathbf{H_e}))\\
\mathbf{p}^{o-end} = \rm{softmax}(\mathbf{v_4^T}\rm{tanh}(\mathbf{W_7} \mathbf{h}_d + \mathbf{W_8}\mathbf{H_e}))
\label{pe}
\end{align}
where $\mathbf{W}_r\in\mathbb{R}^{t\times d}$, $\{\mathbf{W}_i\in\mathbb{R}^{d\times d}\}_{i=1}^8$ and $\{\mathbf{v}_i\in\mathbb{R}^d\}_{i=1}^4$ are learnable parameters, $t$ is the total number of relation types (including a special relation type $\varnothing$ to indicate no triple), $l$ is the sentence length, and $\textbf{H}_e$ is the output of the BERT model.

\subsection{Bipartite Matching Loss}
\label{loss}
The main difficulty of training is to score the predicted triples with respect to the ground truths. It is not proper to apply cross-entropy loss function to measure the difference between two sets, since cross-entropy loss is sensitive to the permutation of the predictions. Inspired by the assigning problem in operation research \cite{kuhn1955hungarian}, we propose a set prediction loss that can produce an optimal bipartite matching between predicted and ground truth triples.

\textbf{Notations.} Let us denote by $\mathbf{Y} = \{\mathbf{Y}_i\}_{i=1}^n$ the set of ground truth triples, and $\hat{\mathbf{Y}} = \{\hat{\mathbf{Y}}_i\}_{i=1}^m$ the set of m predicted triples, where $m$ is larger than $n$. We consider $\mathbf{Y}$ also as a set of size $m$ padded with $\varnothing$ (no triple).  Each element $i$ of the ground truth set can be seen as a $\mathbf{Y}_i = (r_i,s^{start}_i, s^{end}_i,o^{start}_i,o^{end}_i)$,  where $r_i$ is the target relation type (which may be $\varnothing$) and $s^{start}_i, s^{end}_i,o^{start}_i,o^{end}_i$ are the starting or ending indices of subject $s$ or object $o$. Each element $i$ of the set of predicted  triples is denoted as $\hat{\mathbf{Y}}_i =(\mathbf{p}^r_i, \mathbf{p}^{s-start}_i ,\mathbf{p}^{s-end}_i ,\mathbf{p}^{o-start}_i,\mathbf{p}^{o-end}_i)$, which is calculated based on Equation \ref{pr}-\ref{pe}.

\textbf{Loss. } The process of computing bipartite matching loss is divided into two steps: finding an optimal matching and computing the loss function.

To find an optimal matching between the set of ground truth triples $\mathbf{Y}$ and the set of predicted triples $\hat{\mathbf{Y}}$, we search for a permutation of elements $\pi^\star$ with the lowest cost:
\begin{equation}
    \pi^\star = \mathop{\arg \min} \limits_{\pi \in \Pi(m)} \sum \limits_{i=1}^m \mathcal{C}_{match}(\mathbf{Y}_i, \hat{\mathbf{Y}}_{\pi(i)})  
    \label{assignment}
\end{equation}
where $\Pi(m)$ is the space of all m-length permutations.  $\mathcal{C}_{match}(\mathbf{Y}_i, \hat{\mathbf{Y}}_{\pi(i)})$ is a pair-wise matching cost between the ground truth $\mathbf{Y}_i$ and the predicted triple with index $\pi(i)$. By taking into account both the prediction of relation type and the predictions of entity spans, we define $\mathcal{C}_{match}(\mathbf{Y}_i, \hat{\mathbf{Y}}_{\pi(i)})$ as:
\begin{equation}
\begin{split}
     \mathcal{C}_{match}(\mathbf{Y}_i, \hat{\mathbf{Y}}_{\pi(i)})  &= - \mathbbm{1} _{\{r_i \neq \varnothing\}}[\mathbf{p}^r_{\pi(i)}(r_i) \\
    & + \mathbf{p}^{s-start}_{\pi(i)}(s_i^{start})\\
    &+ \mathbf{p}^{s-end}_{\pi(i)}(s_i^{end})\\
    &   + \mathbf{p}^{o-start}_{\pi(i)}(o_i^{start}) \\
    &+ \mathbf{p}^{o-end}_{\pi(i)}(o_i^{end})]
\end{split}
\end{equation}

This optimal assignment $\pi^\star$ is computed in polynomial time ($O(m^3)$) via the Hungarian algorithm \footnote{ \url{https://en.wikipedia.org/wiki/Hungarian_algorithm}}. In detail, we can view the set of ground truth $\mathbf{Y}$ as a set of people in the assignment problem, the set of predicted triples $\mathbf{\hat{Y}}$ as a set of jobs. The cost of assigning $\mathbf{Y}_i$ (the people $i$) with $\mathbf{\hat{Y}_j}$ (the job $j$) is defined as  $\mathcal{C}_{match}(\mathbf{Y}_i, \hat{\mathbf{Y}}_{j})$. The optimal matching with the minimum total cost is easy to be computed via the classical Hungarian algorithm.

The second step is to compute the loss function for all pairs matched in the previous step. We define the loss as:
\begin{equation}
\begin{split}
    \mathcal{L}(\mathbf{Y}, \hat{\mathbf{Y}})  &= \sum \limits_{i=1}^m\{-\log \mathbf{p}^r_{\pi^\star(i)}(r_i) \\
    & + \mathbbm{1} _{\{r_i \neq \varnothing\}}[-\log \mathbf{p}^{s-start}_{\pi^\star(i)}(s_i^{start})\\
    &-  \log \mathbf{p}^{s-end}_{\pi^\star(i)}(s_i^{end})\\
    & - \log \mathbf{p}^{o-start}_{\pi^\star(i)}(o_i^{start}) \\
    & - \log \mathbf{p}^{o-end}_{\pi^\star(i)}(o_i^{end})]\}
\end{split}
\label{loss_equ}
\end{equation}
where $\pi^\star$ is the optimal assignment computed in the first step  (Equation \ref{assignment}).
\section{Experiments}
In this section, we carry out an extensive set of experiments with the aim of answering the following research questions: 
\begin{itemize}
    \item  \textbf{RQ1}: What is the overall performance of the proposed set prediction networks (SPN) in joint entity and relation extraction?
    \item \textbf{RQ2}: How does each design of our model matter? 
    \item \textbf{RQ3}: What is the performance of the proposed model in sentences annotated with different numbers of triples?
    \item \textbf{RQ4}: How does our proposed model adapt to different overlapping patterns?
\end{itemize}
In the remainder of this section, we describe the datasets, experimental setting, and all baselines.

\subsection{Datasets}
We evaluate the proposed method on two widely used joint entity and relation extraction datasets: New York Times (NYT) \cite{riedel2010modeling} and WebNLG \cite{gardent2017creating} \footnote{These datasets are available at \url{https://github.com/xiangrongzeng/copy_re}}. The statistics of these datasets are shown in Table \ref{dataset}.

\textbf{NYT}. This dataset is produced by the distantly supervised method, which automatically aligns Freebase with 1987-2007 New York Times news articles. There are 24 predefined relation types in total. Following previous studies \cite{zheng2017joint,zeng2018extracting}, we ignore the noise in this dataset and treat it as a supervised dataset. Since there are many versions of the NYT dataset, we adopted the preprocessed dataset used in \citet{zeng2018extracting}, which is publicly available. 

\textbf{WebNLG}. This data is originally created for Natural Language Generation (NLG) task. In this dataset, an instance includes a set of triples and several standard sentences written by humans. Every standard sentence contains all triples of this instance. There are 246 predefined relation types in this dataset. 

\textbf{Overlap Between Triples}. According to different overlapping patterns of triples, sentences are split into three categories \cite{zeng2018extracting}: Normal, Entity Pair Overlap (EPO) and Single Entity Overlap (SEO). A sentence belongs to Normal class if none of its triples have overlapped entities. A sentence belongs to EPO class if some of its triples have overlapped entity pairs. And a sentence belongs to SEO class if some of its triples have an overlapped entity and these triples don’t have overlapped entity pair. Note that a sentence
can belongs to both EPO class and SEO class.
\begin{table}[h]
\begin{center}
\scalebox{0.88}{
\begin{tabular}{c|cc|cc}
\hline
\multirow{2}{*}{Category} &
\multicolumn{2}{c|}{NYT} & \multicolumn{2}{c}{WebNLG} \\
\cline{2-5}
                          & Train       & Test      & Train        & Test        \\
                          \hline\hline
Normal                    & 37013       & 3266      & 1596         & 246         \\
EPO                       & 9782        & 978       & 227          & 26          \\
SEO                       & 14735       & 1297      & 3406         & 457         \\ \hline
ALL                       & 56195       & 5000      & 5019         & 703        \\ \hline
\end{tabular}}
\caption{The statistics of NYT and WebNLG.}
\label{dataset}
\end{center}
\end{table}

\begin{table*}[thbp]
\begin{center}
\begin{tabular}{l|ccc|ccc}
\hline
\multirow{2}{*}{Models}               & \multicolumn{3}{c|}{Partial Matching} & \multicolumn{3}{c}{Exact Matching} \\\cline{2-7}

                                      & Precision    & Recall    & F1     & Precision   & Recall   & F1     \\
                                      \hline\hline
NovelTagging \cite{zheng2017joint}     & 62.4         & 31.7      & 42     & -            &  -        &   -     \\
CopyRE-One \cite{zeng2018extracting}   & 59.4         & 53.1      & 56.0   &   -          &    -      &   -     \\
CopyRE-Mul\cite{zeng2018extracting} & 61.0         & 56.6      & 58.7   &  -           &     -     &    -    \\
GraphRel-1p \cite{fu2019graphrel}          & 62.9         & 57.3      & 60.0   & -            & -         &    -    \\
GraphRel-2p \cite{fu2019graphrel}            & 63.9         & 60.0      & 61.9   &        -     &      -    &      -  \\
CopyRRL \cite{zeng2019learning}          & 77.9         & 67.2      & 72.1   &           -  &    -      &   -     \\
WDec\cite{nayak2019ptrnetdecoding}     &   -           &  -         &  -                    &88.1         & 76.1      & 81.7        \\
PNDec\cite{nayak2019ptrnetdecoding} &   -           &  -         &  -                    &80.6         & 77.3      & 78.9        \\
CopyMTL-One \cite{zeng2020copymtl} &-&-&-&72.7 &69.2 &70.9 \\
CopyMTL-Mul \cite{zeng2020copymtl} &-   &- &- &  75.7 &68.7 &72.0 \\
Attention as Relation \cite{ijcai2020-524} &-&-&-&88.1& 78.5& 83.0\\
CasRel \cite{wei-etal-2020-novel}                        & 89.7         & 89.5      & 89.6   & 90.1$ ^\dag$       & 88.5$ ^\dag$    & 89.3 $ ^\dag$\\
\hline
    SPN (Ours)                                  & \textbf{93.3}         & \textbf{91.7}      & \textbf{92.5}   &  \textbf{92.5}           &   \textbf{92.2}       & \textbf{92.3}    \\  
\hline
\end{tabular}
\caption{Precision (\%) , Recall (\%)  and F1 score (\%) of our proposed SPN and state-of-the-art mehtods on the NYT test set. $\dag$ indicates that the result is reproduced by us.}
\label{NYT_result}
\end{center}
\end{table*}

\begin{table}[thbp]
\begin{center}
\begin{tabular}{l|ccc}
\hline
Models  & Precision    & Recall    & F1  \\
                                      \hline\hline
NovelTagging     & 52.5         & 19.3      & 28.3       \\
CopyRE-One   & 32.2         &28.9      & 30.5    \\
CopyRE-Mul & 37.7         &36.4     & 37.1    \\
GraphRel-1p        & 42.3         & 39.2     & 40.7    \\
GraphRel-2p           & 44.7         & 41.1     & 42.9  \\
CopyRRL       &63.3         & 59.9      & 61.6  \\
CasRel                       & \textbf{93.4}         & 90.1      & 91.8 \\
\hline
SPN (Ours) & 93.1         & \textbf{93.6}      & \textbf{93.4} \\  
\hline
\end{tabular}
\caption{Precision (\%) , Recall (\%)  and F1 score (\%) of our proposed SPN and state-of-the-art mehtods on the WebNLG test set.}
\label{WebNLG_result}
\end{center}
\end{table}

\subsection{Evaluation Metrics}
We adopt standard micro-F1 to evaluate the performance. A triple is regarded as correct if the relation type and the two corresponding entities are all correct. Notably, there are two ways to judge whether the extracted entities are correct. One is Partial Matching, where the extracted entities are regarded as correct if the predictions of subject and object are the same as the head words of the ground truth. Since the head word of entity is not annotated in most situations, the last word of entity is treated as the head word. The other is Exact Matching. In this way, only if the predictions of subject and object are identical to the ground truth, the extracted entities are treated as correct. Note that the training data under partial matching is different from the one under exact matching. Under partial matching, only head words of entities are annotated, while the whole entities  are annotated under exact matching.
\subsection{Implementation Details}
To conduct a fair comparison, we use the cased base version of BERT in our experiments, which contains 110M parameters. The initial learning rate of BERT is set to 0.00001, and the initial learning of the non-autoregressive decoder is set to 0.00002. The number of stacked bidirectional transformer blocks in non-autoregressive decoder is set to 3. We use the dropout strategy to mitigate overfitting, the dropout rate is set to 0.1. Meanwhile, We apply gradient clipping to prevent exploding gradients. The set prediction networks are trained by minimizing the loss function defined in Equation \ref{loss_equ} through stochastic gradient descent over shuffled mini-batch with the AdamW update rule  \cite{loshchilov2017decoupled}. All experiments are conducted with an NVIDIA GeForce RTX 2080 Ti.

\begin{table*}[thbp]
\begin{center}
\begin{tabular}{c|c|ccc|ccc}
\hline
\multirow{2}{*}{Models}    & \multirow{2}{*}{Element} & \multicolumn{3}{c|}{NYT}   & \multicolumn{3}{c}{WebNLG} \\
\cline{3-8}
                           &                          & Precsoion & Recall & F1   & Precision  & Recall & F1   \\
                           \hline \hline
\multirow{3}{*}{CasRel \cite{wei-etal-2020-novel}}    & ($s$,$o$)                    & 89.2      & 90.1   & 89.7 & \textbf{95.3}       & 91.7   & 93.5 \\
                           & $r$                        & 96.0     & 93.8   & 94.9 & \textbf{96.6}       & 91.5   & 94.0 \\
                           & Overall   & 89.7         & 89.5      & 89.6  &  \textbf{93.4}         & 90.1      & 91.8 \\
                           \hline
\multirow{3}{*}{SPN (Ours)} & $(s,o)$                    & \textbf{93.2}      & \textbf{92.7}   & \textbf{92.9} & 95.0       & \textbf{95.4}   & \textbf{95.2} \\
                           & $r$                        & \textbf{96.3}     & \textbf{95.7}   & \textbf{96.0} & 95.2       & \textbf{ 95.7}   & \textbf{95.4}\\
                           & Overall & \textbf{93.3}         & \textbf{91.7}      & \textbf{92.5} & 93.1         & \textbf{93.6}      & \textbf{93.4}\\
                           \hline
\end{tabular}
\end{center}
\caption{Results on extracting elements of relational triples.}
\label{detail_result}
\end{table*}

\subsection{Baselines}
The following state-of-the-art (SoTA) models have been compared in the experiments.
\begin{itemize}
\item NovelTagging \cite{zheng2017joint} introduces a novel tagging scheme that transforms the joint entity and relation extraction task into a sequence labeling problem.
\item CopyRE \cite{zeng2018extracting} is a seq2seq based model with copy mechanism, which can effectively extract overlapping triples.
\item GraphRel \cite{fu2019graphrel} is a two phases model based on graph convolutional networks (GCN), where a relation-weighted GCN is utilized to
model the interaction between entities and relations.
\item CopyRRL \cite{zeng2019learning} combines the reinforcement
learning with a seq2seq model to automatically learn the
extraction order of triples. In such a way, the interactions among
triples can be considered.
\item CopyMTL \cite{zeng2020copymtl} is a multi-task learning
framework, where conditional random field is used to identify entities, and
a seq2seq model is adopted to extract relational triples.
\item WDec \cite{nayak2019ptrnetdecoding} fuses a seq2seq model with a new representation scheme, which enables the decoder to generate one word at a and can handle full entity names of different length and overlapping entities. 
\item PNDec \cite{nayak2019ptrnetdecoding} is a modification of seq2seq model. Pointer networks are used in the decoding framework to identify the entities in the sentence using their start and end locations.
\item Attention as Relation \cite{ijcai2020-524} contains
a conditional random field based entity extraction module and a supervised multi-head self attention based relation detection module. 
\item CasRel \cite{wei-etal-2020-novel} is a novel cascade binary tagging
framework, where all possible subjects are identified in the first stage, and then for each identified subject, all possible relations and the corresponding objects are simultaneously identified by a relation specific tagger. This work achieves the SoTA results.
\end{itemize}

Note that \citet{li2020downstream} claimed that they have achieved the SoTA results in WebNLG, but we find that the model they proposed is only designed for relation classification. Therefore, we do not compare our method with REDN \cite{li2020downstream}.

\subsection{Main Results}

To start, we address the research question \textbf{RQ1}. Table \ref{NYT_result} and Table \ref{WebNLG_result} show the results of our model against baselines on two benchmark datasets. Overall, our proposed model significantly outperforms baselines on these datasets.

In NYT dataset, Our proposed model outperforms all the baselines in both partial matching and exact matching and achieves 2.9\% and 3.0\% improvements in F1 score respectively over the current SoTA method \cite{wei-etal-2020-novel}. Combined with Table \ref{detail_result}, we find that our proposed model outperforms the SoTA model \cite{wei-etal-2020-novel} in both entity pair extraction and relation type extraction, demonstrating the effectiveness of our proposed model. We also find that there is an obvious gap between the F1 score on relation type extraction and entity pair extraction, but a trivial gap between entity pair extraction and overall extraction. It reveals that entity pair extraction is the main bottleneck of joint entity and relation extraction in NYT dataset. 

In WebNLG dataset, since there is no unified version of data under exact matching, we only compare our proposed networks with baselines under partial matching metric. In general, our proposed model achieves the best F1 score under partial matching, which is 93.4\%. There is 1.6\% improvement compared with the result of the SoTA model \cite{wei-etal-2020-novel}, which is 91.8\%. Compared with the SoTA model, We find our model to be more balanced in terms of precision and recall. A further analysis combined with Table \ref{detail_result} shows that our proposed model exhibits balanced results of precision and recall in both entity pair extraction and relation type extraction, while there is an obvious gap between precision and recall in the results of CasRel \cite{wei-etal-2020-novel}. We conjecture that this is largely due to that the number of triple queries is set to be significantly larger than the typical number of triples in a sentence, which ensures that enough triples can be recalled. In addition, we observe that the results of entity pair extraction and relation type extraction are much higher than that of joint entity and relation extraction, which means that the accurate combination of entity pair extraction and relation type extraction is the key to improve the joint entity and relation extraction.
\begin{table}[t]
\scalebox{0.96}{
\begin{tabular}{l|ccc}
\hline
                                                                                            & Precision & Recall & F1   \\
                                                                                            \hline \hline
\begin{tabular}[c]{@{}l@{}}Bipartite Matching Loss, \\ Num of Decoder Layers = 3\end{tabular} & \textbf{93.3}      & \textbf{91.7}   & \textbf{92.5} \\
\hline
\begin{tabular}[c]{@{}l@{}}Bipartite Matching Loss, \\ Num of Decoder Layers = 2\end{tabular} & 92.7      & 91.3   & 92.0 \\
\hline
\begin{tabular}[c]{@{}l@{}}Bipartite Matching Loss, \\ Num of Decoder Layers = 1\end{tabular}  & 91.9      & 90.9   & 91.4 \\
\hline
\begin{tabular}[c]{@{}l@{}}Cross-Entropy Loss,\\ Num of Decoder Layers = 3\end{tabular}   & 87.2     & 80.9   & 84.0 \\
\hline
\end{tabular}}
\caption{Results of ablation studies on NYT dataset.}
\label{oblation}
\end{table}

% \begin{figure}[t]
%   	\begin{center} \includegraphics*[clip=true,width=0.45\textwidth,height=0.22\textheight]{graph/oblation.pdf}
%   	\caption{F1 score of our proposed model with different number of decoder layers.}
%   	\label{oblation}
%   \end{center}
% \end{figure}

 \begin{table*}[thbp]
\begin{center}
\begin{tabular}{l|ccccc|ccccc}
\hline
\multirow{2}{*}{Models} & \multicolumn{5}{c|}{NYT}          & \multicolumn{5}{c}{WebNLG}       \\\cline{2-11}
                        & N=1  & N=2  & N=3  & N=4  & N$\geq$5 & N=1  & N=2  & N=3  & N=4  & N$\geq$5 \\
                        \hline \hline
CopyRE-One \cite{zeng2018extracting}             & 66.6 & 52.6 & 49.7 & 48.7 & 20.3 & 65.2 & 33.0 & 22.2 & 14.2 & 13.2 \\
CopyRE-Mul \cite{zeng2018extracting}         & 67.1 & 58.6 & 52.0 & 53.6 & 30.0 & 59.2 & 42.5 & 31.7 & 24.2 & 30.0 \\
GraphRel-1p \cite{fu2019graphrel} & 69.1 & 59.5 & 54.4 & 53.9 & 37.5 & 63.8 & 46.3 & 34.7 & 30.8 & 29.4 \\
GraphRel-2p \cite{fu2019graphrel}  & 71.0 & 61.5 & 57.4 & 55.1 & 41.1 & 66.0 & 48.3 & 37.0 & 32.1 & 32.1 \\
CopyRRL \cite{zeng2019learning} & 71.7 & 72.6 & 72.5 & 77.9 & 45.9 & 63.4 & 62.2 & 64.4 & 57.2 & 55.7 \\
CasRel \cite{wei-etal-2020-novel} & 88.2 & 90.3 & 91.9 & 94.2 & 83.7 & 89.3 & 90.8 & 94.2 & 92.4 & 90.9 \\ \hline
SPN (Ours)              & \textbf{90.9} & \textbf{93.4} & \textbf{94.2} & \textbf{95.5} & \textbf{90.6} & \textbf{89.5} & \textbf{91.3} & \textbf{96.4} & \textbf{94.7} & \textbf{93.8}\\
\hline
\end{tabular}
\caption{Partial matching F1 score of conducting extraction in sentences that contains different numbers of triples. We divide the sentences of the test sets into 5 sub-classes. Each class contains sentences that have 1,2,3,4 or $\geq$5 triples.}
\label{num_reulst}
\end{center}
\end{table*}

\subsection{Ablation Studies}
Next, we turn to the research question \textbf{RQ2}. We conduct ablation studies to investigate the importance of the non-autoregressive decoder and the bipartite matching loss. To evaluate the importance of the non-autoregressive decoder, we change the number of decoder layers. To evaluate the importance of bipartite matching loss, we replacing bipartite matching loss with cross-entropy loss.

Table \ref{oblation} shows the results. We find that increasing the number of layers of the non-autoregressive decoder can achieve better results. When the number of decoder layers is set to 1, 2, and 3, the best results are 91.4\%, 92.0\% and 92.15\%, respectively. We conjecture that this is largely due to that with the deepening of the non-autoregressive decoder layers,  more multi-head self attention modules allow for better modeling of relationships between triple queries, and more multi-head inter attention modules allow for more complete integration of sentence information into triple queries. 

Compared our proposed bipartite matching loss with widely used cross-entropy loss, we find that there is 8.9\% improvement, which indicates bipartite matching loss is very suitable for joint entity and relation extraction. We hypothesize that in order to adopt to cross-entropy loss, the model is required to consider the generative order of triples, which places an unnecessary burden on the model.

\subsection{Detailed Results on Sentences with Different Number of Triples}
In this section, we answer the research question \textbf{RQ3}. To do this, we compare the models' ability of extracting relational facts from sentences annotated with different numbers of triples. We divide the sentences in test sets into 5 sub-classes. Each class contains sentences that have 1, 2, 3, 4 or $\geq$5 triples. The results are shown in Table \ref{num_reulst}. In general, our proposed model achieves the best results in all sub-classes. Besides, we observe that though the our proposed model gains considerable improvements on all five sub-classes compared to the SoTA model \cite{wei-etal-2020-novel}, the greatest improvement of F1 score on the two datasets both come from the most difficult sub-class (N$\geq$5), in particular, there is 6.9\% improvement in the NYT data. Such results indicate that our proposed model is more suitable for complicated scenarios than the current SoTA model.
\subsection{Detailed Results on Different Overlapping Patterns}
\begin{figure}[t]
  	\begin{center} \includegraphics*[clip=true,width=0.47\textwidth,height=0.22\textheight]{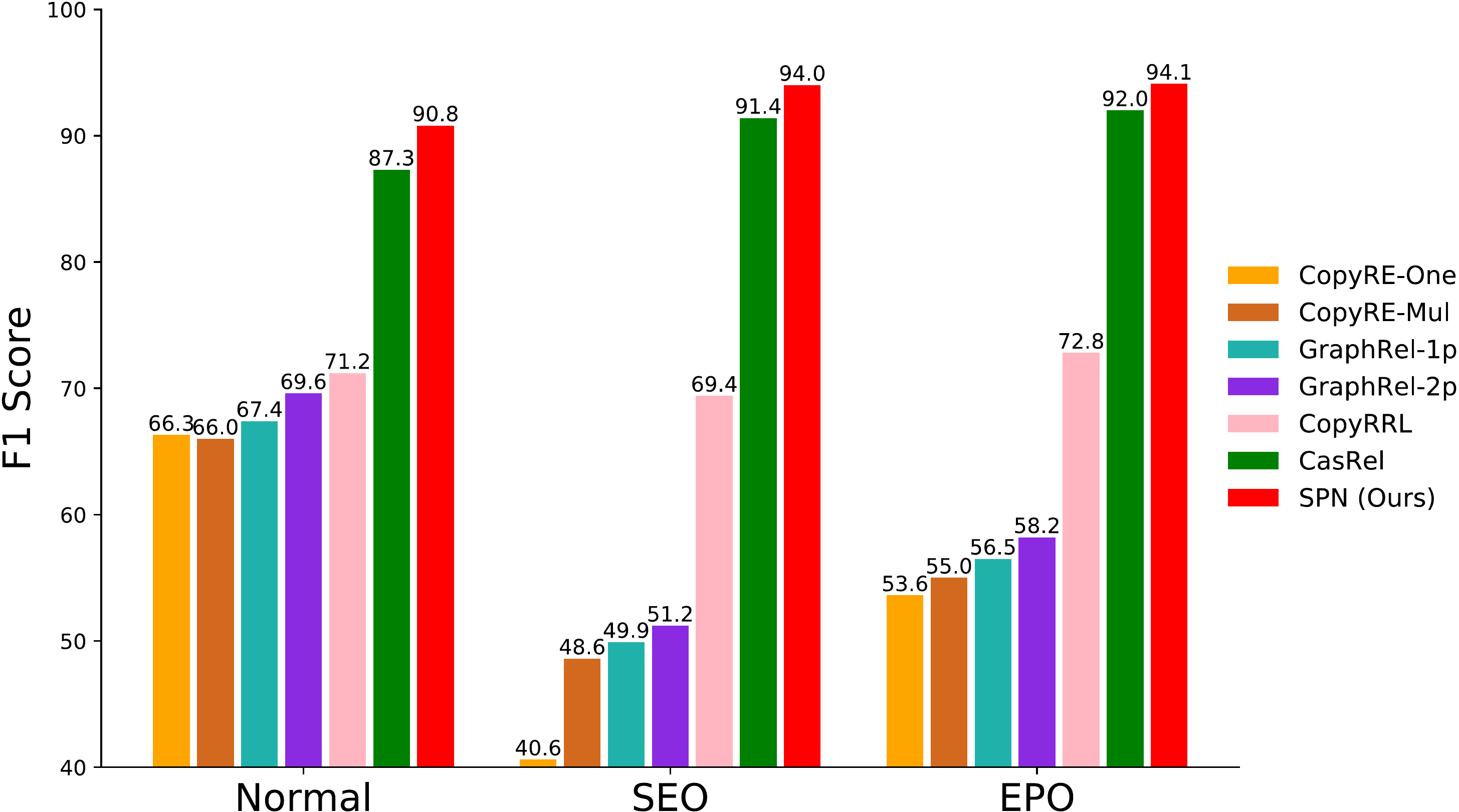}
  	\caption{F1 score of extracting relational triples from sentences with different overlapping pattern in NYT dataset.} \label{overlap}
  \end{center}
  \end{figure}

Finally, to address  the research question \textbf{RQ4}, we conduct further experiments on NYT dataset to verify the ability of our proposed model in handling the overlapping problem.  Figure \ref{overlap} shows the results of our proposed model and baseline models in Normal, Single Entity Overlap (SEO) and Entity Pair Overlap (EPO) classes. 

Overall, our proposed model performs better than all baseline models in all three classes.  In details, there are 3.5\%, 2.6\% and 2.1\% improvements compared with the SoTA model \cite{wei-etal-2020-novel} in Normal, SEO and EPO classes, respectively. Such results show that our proposed model is very effective in handling the overlapping problem, which is widely existed in joint entity and relation extraction. In addition,  We also observe that the performance of all seq2seq based models, such as CopyRE \cite{zeng2018extracting} and CopyRRL\cite{zeng2019learning}, on Normal, EPO and SEO presents a decreasing trend, which indicates that the Normal class presents a relatively easiest pattern while EPO and SEO classes are the relatively harder ones for these seq2seq based models to extract. In contrast, our 
proposed model attains consistently strong performance over all three overlapping patterns.

% \begin{figure*}[t]
%   	\begin{center} \includegraphics*[clip=true,width=1\textwidth,height=0.2\textheight]{graph/overlap.pdf}
%   	\caption{F1 score of extracting relational triplets from sentences with different overlapping pattern.
%   	} \label{fig}
%   \end{center}
%   \end{figure*}

\section{Conclusion and Future Work}
In this paper, we introduce set prediction networks for joint entity and relation extraction. Compared with previous seq2seq based models, We formulate the  joint entity and relation extraction task as a set prediction problem. In such a way, the extraction model will be relieved of predicting the extraction order of multiple triples. To solve the set prediction problem, We combine non-autoregressive parallel decoding with bipartite matching loss function. We conduct extensive experiments on two widely used datasets to validate the effectiveness of the proposed set prediction networks. Experimental results show that our proposed networks outperforms state-of-the-art baselines over different scenarios. This challenging task is far from being solved. We find that relation types exhibit an imbalanced or long-tailed distribution in NYT dataset and WebNLG dataset. Our future work will concentrate on how to combine cost-sensitive learning with the proposed set prediction networks.

\bibliography{aaai21}

\section{Hungarian Algorithm}

The Hungarian method is a combinatorial optimization algorithm that solves the assignment problem in polynomial time. The Hungarian algorithm consists of the four steps below. The first two steps are executed once, while Steps 3 and 4 are repeated until an optimal assignment is found. The input of the algorithm is an $m \times m$ by m square matrix, called cost matrix.
\\

\noindent \textbf{Step 1}: Subtract row minima. \\

\noindent For each row, find the lowest element and subtract it from each element in that row.\\

\noindent \textbf{Step 2}: Subtract column minima.  \\

\noindent Similarly, for each column, find the lowest element and subtract it from each element in that column. \\

\noindent \textbf{Step 3}: Cover all zeros with a minimum number of lines.\\

\noindent Cover all zeros in the resulting matrix using a minimum number of horizontal and vertical lines. If $m$ lines are required, an optimal assignment exists among the zeros. The algorithm stops. \\

\noindent If less than $m$ lines are required, continue with Step 4.\\

\noindent \textbf{Step 4}: Create additional zeros
\\

\noindent  Find the smallest element (call it k) that is not covered by a line in Step 3. Subtract k from all uncovered elements, and add k to all elements that are covered twice.

\section{An Example of Bipartite Matching Loss}
We use an example to illustrate how bipartite matching loss function works. In this example, we assume that there are only three relation types that we are interested in. These three relation types are \texttt{leader\_name}, \texttt{located\_in} and \texttt{capital\_Of}. As shown in this paper, a special relation type $\varnothing$ is also required. Furthermore, we assume that the non-autoregressive decoder generate 3 triples for each sentence, in other words, $m$ is set to 3. The sentence, the ground-truths 
and the predictions are presented as follows:

\subsection{Sentence}
% \noindent \textbf{Sentence}:\\
\texttt{Aarhus airport serves the city of} \texttt{Aarhus ,}  \texttt{which is led by Jacob Bundsgaard .}
\\

\subsection{Ground-truths}
% \noindent \textbf{Ground-Truths}:\\
\texttt{(Aarhus, leader\_name, Jacob Bundsgaard)} \\
\texttt{(Aarhus Airport, located\_in, Aarhus)} \\
$\varnothing$ (padding a no triple to match $m$) \\

\noindent The indices of \texttt{Aarhus Airport}, \texttt{Aarbus} and \texttt{Jacob Bundsgaard} are $(0, 1)$, $(6, 6)$ and $(12, 13)$ respectively. We also convert each relations type to a unique integer. In detail, \texttt{leader\_name}, \texttt{located\_in},  \texttt{capital\_Of} and  $\varnothing$ are converted to 0, 1, 2, 3 respectively. In such a way, ground-truths are represented as: \\
\begin{equation*}
\begin{split}
\mathbf{Y}_0 &= \{0, 6, 6, 12, 13\} \\
\mathbf{Y}_1 &= \{1, 0, 1, 6, 6\} \\
\mathbf{Y}_2 & = \{3\}
\end{split}
\end{equation*}

\subsection{Predictions}

We assume that the outputs of the set prediction networks are as follow: 

% \noindent \textbf{Predictions}:\\
\noindent $\hat{\mathbf{Y}}_0 = \{$ \\
\indent \quad \quad $p_0^r$ \quad   \quad\, :$\ (0.1, 0.3, 0.4, 0.2)$
\\
\indent \quad \quad   $p_0^{s-start}: (0.9, 0.1, 0, 0, 0, 0,0, 0, 0, 0, 0, 0, 0, 0, 0)$\\
% \indent \quad \quad \quad\quad \quad\quad \quad $$ \\
\indent \quad \quad  $p_0^{s-end}\quad\!\!\! : (0.2, 0.8, 0, 0, 0, 0, 0, 0, 0, 0, 0, 0, 0, 0, 0)$\\
\indent \quad \quad   $p_0^{o-start}: (0.1, 0.1, 0, 0, 0, 0, 0.7, 0, 0, 0, 0, 0, 0.1, 0, 0)$\\
\indent \quad \quad $p_0^{o-end}\quad\!\!\! : (0, 0.1, 0, 0, 0, 0, 0.6, 0, 0, 0, 0, 0, 0.1, 0.2, 0)$\\
% \indent \quad \quad \quad\quad \quad\quad \quad $0, 0.025, 0.025, 0.075, 0, 0.1, 0)$ \\
\indent \ \ \ \ \ \ \  $\}$ \\

\noindent $\hat{\mathbf{Y}}_1 = \{$ \\
\indent \quad \quad  $p_1^r$ \quad   \quad\, :$\ (0.5, 0.25, 0.15, 0.1)$
\\
\indent \quad \quad   $p_1^{s-start}: (0.1, 0, 0, 0, 0, 0, 0.8, 0,0,0,0,0,0.1,0,0)$\\
% \indent \quad \quad \quad\quad \quad\quad \quad $0.025, 0.025, 0, 0.025, 0.075, 0, 0, 0)$ \\
\indent \quad \quad  $p_1^{s-end}\quad\!\!\! : (0.2, 0.3, 0, 0, 0, 0, 0.5, 0,0,0,0,0,0,0,0)$\\
% \indent \quad \quad \quad\quad \quad\quad \quad $0, 0.025, 0.025, 0.075, 0, 0.1, 0)$ \\
\indent \quad \quad   $p_1^{o-start}: (0.2, 0.1, 0, 0, 0, 0, 0.2, 0,0,0,0,0,0.5,0,0)$\\
% \indent \quad \quad \quad\quad \quad\quad \quad $0.025, 0.025, 0, 0.025, 0.075, 0, 0.1, 0)$ \\
\indent \quad \quad  $p_1^{o-end}\quad\!\!\! : (0, 0.4, 0, 0, 0, 0, 0.3, 0,0,0,0,0,0,0.3,0)$\\
% \indent \quad \quad \quad\quad \quad\quad \quad $0, 0.025, 0.025, 0.075, 0, 0.1, 0)$ \\
\indent \ \ \ \ \ \ \  $\}$ \\

\noindent $\hat{\mathbf{Y}}_2 = \{$ \\
\indent \quad \quad $p_2^r$ \quad   \quad\, :$\ (0.1, 0.3, 0.4, 0.2)$
\\
\indent \quad \quad $p_2^{s-start}: (0.4, 0, 0, 0, 0, 0,0.5, 0,0,0,0,0,0.1,0,0)$\\
% \indent \quad \quad \quad\quad \quad\quad \quad $0.025, 0.025, 0, 0.025, 0.075, 0, 0.1, 0)$ \\
\indent \quad \quad $p_2^{s-end}\quad\!\!\! : (0.1, 0.4, 0, 0, 0, 0.5, 0, 0, 0, 0, 0, 0, 0, 0, 0)$\\
% \indent \quad \quad \quad\quad \quad\quad \quad $0, 0.025, 0.025, 0.075, 0, 0.1, 0)$ \\
\indent \quad \quad $p_2^{o-start}: (0.3, 0.1, 0, 0, 0, 0, 0, 0, 0, 0, 0, 0, 0.7, 0, 0)$\\
% \indent \quad \quad \quad\quad \quad\quad \quad $0.025, 0.025, 0, 0.025, 0.075, 0, 0.1, 0)$ \\
\indent \quad \quad $p_2^{o-end}\quad\!\!\! : (0, 0.2, 0, 0, 0, 0, 0.4, 0, 0, 0, 0,0 ,0,0.4,0)$\\
% \indent \quad \quad \quad\quad \quad\quad \quad $0, 0.025, 0.025, 0.075, 0, 0.1, 0)$ \\
\indent \ \ \ \ \ \ \  $\}$
\subsection{Bipartite Matching Loss}
The process of computing bipartite matching loss is divided into two steps: finding an optimal matching and computing the loss function.\\

\noindent \textbf{(1). Finding an optimal matching} \\
\noindent We search for a permutation of elements $\pi^\star$ with the lowest cost to find an optimal matching between the set of ground truth triples $\mathbf{Y}$ and the set of predicted triples $\hat{\mathbf{Y}}$:
\begin{equation}
    \pi^\star = \mathop{\arg \min} \limits_{\pi \in \Pi(m)} \sum \limits_{i=1}^m \mathcal{C}_{match}(\mathbf{Y}_i, \hat{\mathbf{Y}}_{\pi(i)})  
    \label{assignment}
\end{equation}
To do this, we need to determine the cost matrix, which is the input of Hungarian algorithm. The (i, j)-entry of cost matrix is $\mathcal{C}_{match}(\mathbf{Y}_i, \hat{\mathbf{Y}}_{j})$, where $\mathcal{C}_{match}(\mathbf{Y}_i, \hat{\mathbf{Y}}_{j})$ is define as:
\begin{equation*}
\begin{split}
     \mathcal{C}_{match}(\mathbf{Y}_i, \hat{\mathbf{Y}}_{j})  &= - \mathbbm{1} _{\{r_i \neq \varnothing\}}[\mathbf{p}^r_{j}(r_i) \\
    & + \mathbf{p}^{s-start}_{j}(s_i^{start})\\
    &+ \mathbf{p}^{s-end}_{j}(s_i^{end})\\
    &   + \mathbf{p}^{o-start}_{j}(o_i^{start}) \\
    &+ \mathbf{p}^{o-end}_{j}(o_i^{end})]
\end{split}
\end{equation*}
Each entry of cost matrix is computed as follows: 
\begin{equation*}
\begin{split}
\mathcal{C}_{match}(\mathbf{Y}_0, \hat{\mathbf{Y}}_{0}) &= -[0.1+0+0+0.1+0.2] = -0.4\\
\mathcal{C}_{match}(\mathbf{Y}_0, \hat{\mathbf{Y}}_{1}) &= -[0.5+0.8+0.5+0.5+0.3] = -2.6\\
\mathcal{C}_{match}(\mathbf{Y}_0, \hat{\mathbf{Y}}_{2}) &= -[0.1+0.5+0.5+0.7+0.3] = -2.1\\
\mathcal{C}_{match}(\mathbf{Y}_1, \hat{\mathbf{Y}}_{0}) &= -[0.3+0.9+0.8+0.7+0.6] = -3.3\\
\mathcal{C}_{match}(\mathbf{Y}_1, \hat{\mathbf{Y}}_{1}) &= -[0.25+0.1+0.3+0.2+0.3] = -1.15\\
\mathcal{C}_{match}(\mathbf{Y}_1, \hat{\mathbf{Y}}_{2}) &= -[0.3+0.4+0.4+0+0.4] = -1.5\\
\mathcal{C}_{match}(\mathbf{Y}_2, \hat{\mathbf{Y}}_{0}) &= 0\\
\mathcal{C}_{match}(\mathbf{Y}_2, \hat{\mathbf{Y}}_{1}) &= 0\\
\mathcal{C}_{match}(\mathbf{Y}_2, \hat{\mathbf{Y}}_{2}) &= 0\\
\end{split}
\end{equation*}
The cost matrix is denoted as follows:
\begin{center}
$\mathcal{C} = $
\begin{blockarray}{cccc}
&
$\hat{\mathbf{Y}}_{0}$ & $\hat{\mathbf{Y}}_{1}$ & $\hat{\mathbf{Y}}_{2}$\\
\begin{block}{c(ccc)}
  $\mathbf{Y}_0$ &-0.4 & -2.6 & -2.1 \\
 $\mathbf{Y}_1$ & -3.3& -1.15 & -1.5 \\
  $\mathbf{Y}_2$ &0 & 0 &  0 &  \\
\end{block}
\end{blockarray}
\end{center}
Via Hugarian algorithm, the optimal assignment  $\pi^\star$  corresponding to the cost matrix is 
\[\rm{ground-truths: }[0,1,2]\longrightarrow\rm{predictions: }[1, 0, 2]\]
which has the minimum total cost (-5.9 = (-2.6)+(-3.3)+0).\\

\noindent \textbf{(2). Computing  the  loss  function based on the optimal assignment.} \\
Recall that the loss is defined as:
\begin{equation*}
\begin{split}
       \mathcal{L}(\mathbf{Y}, \hat{\mathbf{Y}})  &= \sum \limits_{i=1}^m\{-\log \mathbf{p}^r_{\pi^\star(i)}(r_i) \\
    & + \mathbbm{1} _{\{r_i \neq \varnothing\}}[-\log \mathbf{p}^{s-start}_{\pi^\star(i)}(s_i^{start})\\
    &-  \log \mathbf{p}^{s-end}_{\pi^\star(i)}(s_i^{end})\\
    & - \log \mathbf{p}^{o-start}_{\pi^\star(i)}(o_i^{start}) \\
    & - \log \mathbf{p}^{o-end}_{\pi^\star(i)}(o_i^{end})]\}
\end{split}
\end{equation*}
Based on the optimal assignment $\pi^\star$, the value of loss function is:
\begin{equation*}
\begin{split}
    \mathcal{L}(\mathbf{Y}, \hat{\mathbf{Y}})  = &\{-\log \mathbf{p}^r_{1}(0) -\log \mathbf{p}^{s-start}_{1}(6) -  \log \mathbf{p}^{s-end}_{1}(6)\\
    & - \log \mathbf{p}^{o-start}_{1}(12) - \log \mathbf{p}^{o-end}_{1}(13)\} \\
    & + \{-\log \mathbf{p}^r_{0}(1) -\log \mathbf{p}^{s-start}_{0}(0) -  \log \mathbf{p}^{s-end}_{0}(1)\\
    & - \log \mathbf{p}^{o-start}_{0}(6) - \log \mathbf{p}^{o-end}_{0}(6)\}  \\
    & + \{-\log \mathbf{p}^r_{2}(3)\} \\
    =& \{-\log(0.5)-\log(0.8)-\log(0.5)-\log(0.5)\\
    & -\log(0.3)\} +\{-\log(0.3)-\log(0.9)-\log(0.8)\\
    & -\log(0.7)-\log(0.6)\} + \{-\log(0.2)\}\\
    =&7.52
\end{split}
\end{equation*}
\end{document}